\documentclass[letterpaper,journal]{IEEEtran}
\usepackage{amsmath,amsfonts}
\usepackage{amssymb,mathtools}
\usepackage{bm} %
\usepackage{physics}  %
\usepackage{booktabs}      %
\usepackage{tabularx}

\newcommand{\R}{\mathbb{R}}

\newcommand{\Qddp}{\mathcal{Q}}  %

\usepackage{algorithmic}
\usepackage{algorithm}
\usepackage{float}
\usepackage{array}
\usepackage[caption=false,font=footnotesize]{subfig}
\usepackage{stfloats}
\usepackage{url}
\usepackage{graphicx}
\newcommand{\rev}[1]{#1}
\usepackage{cite}
\usepackage[hidelinks]{hyperref}
\begin{document}

\title{\rev{Online} {iLQR-MPC with Whole-Body Control for Robust Humanoid Locomotion}}
\author{Xiang Li, Linlin Ou, Ruihan Wu, and Xinyi Yu$^{*}$%
\thanks{The authors are with Zhejiang University of Technology, Hangzhou, China.}%
\thanks{$^{*}$Corresponding author: Xinyi Yu (e-mail: yuxy@zjut.edu.cn).}%
\thanks{Experiment videos: \protect\url{https://anonymous.4open.science/r/ilqr-wbc-humanoid/}}}%

\maketitle

\begin{abstract}
Humanoid locomotion requires deciding where to step and how to distribute contact forces. However, a disturbance can invalidate the nominal plan, requiring coordinated revision under constraints that switch with every footfall. To solve this online, the proposed approach employs the single-rigid-body model, which retains attitude dynamics and the full contact wrench. An iterative linear-quadratic-regulator MPC (iLQR-MPC) then jointly optimizes contact forces and footstep locations in a single receding-horizon problem, without a separate step planner. Specifically, contact and motion constraints are incorporated into the backward pass through augmented-Lagrangian and relaxed-barrier penalties. A whole-body controller subsequently maps the plan to joint torques at $500\,\mathrm{Hz}$ while preserving contact feasibility. The model-based formulation makes disturbance responses interpretable through explicit costs and constraints, and requires no platform-specific policy training. In MuJoCo simulations, the iLQR-MPC replans at $100\,\mathrm{Hz}$ (mean solve time $8.9\,\mathrm{ms}$); the Unitree G1 sustains steady-state walking and recovers from impulsive $245\,\mathrm{N}$ forward and $175\,\mathrm{N}$ lateral pushes.
\end{abstract}


\section{Introduction}
\IEEEPARstart{H}{umanoid} robots must maintain dynamic balance, track footholds, and respect hard constraints (joint limits, torque bounds, and friction cones) on a floating base with dozens of degrees of freedom (DoF) and hybrid dynamics that switch at every contact event. The dominant paradigm pairs a simplified-model MPC with a whole-body controller (WBC), from the Atlas humanoid~\cite{kuindersma2016optimization} to recent quadrupeds~\cite{bellicoso2018dynamic}; a recent survey identifies it as the standard framework~\cite{gu2025humanoid}. Reinforcement learning offers an alternative through sim-to-real transfer~\cite{radosavovic2024real}, but requires per-platform retraining and gives no formal guarantee on constraint satisfaction.

Within model-based control, competing tasks must be resolved. A weighted-sum objective cannot guarantee safety-critical constraints when they conflict with secondary goals; for a walking humanoid, a violation means a fall. Task-priority control instead enforces tasks in priority order through null-space projection~\cite{sentis2005control,sentis2010compliant}, extended to inequality constraints via hierarchical quadratic programming (HQP)~\cite{escande2014hierarchical} and validated on torque-controlled humanoids~\cite{herzog2016momentum}. Because WBC acts on the instantaneous state alone, it is reactive; MPC adds lookahead by planning contact forces over a receding horizon, but full-order nonlinear MPC is intractable at real-time rates. The standard approach therefore pairs a reduced-model MPC at 50--100\,Hz, which plans ground reaction forces and CoM trajectories, with a WBC at 500--1000\,Hz mapping the plan to joint torques. Linear MPC based on the zero-moment point (ZMP)~\cite{wieber2006trajectory} and nonlinear contact-implicit MPC~\cite{neunert2018whole} have been deployed this way.

External pushes test these constraint margins. Push recovery uses three strategies~\cite{koolen2012capturability}: ankle (ZMP modulation), hip (angular-momentum regulation), and stepping (footstep adjustment). Stepping is the last recourse because it redefines the support base~\cite{pratt2006capture}. Online gait adaptation has followed two directions. The first modifies step \emph{timing}~\cite{khadiv2020walking,li2025adapting}, relying on reduced-order dynamics, chiefly the linear inverted pendulum, for tractability. The second promotes footstep location to a state variable within the MPC, coupled with a foot-velocity control, so footholds and contact forces are co-optimized: Romualdi et al.~\cite{romualdi2022online} introduced this on the position-controlled iCub, and Ding et al.~\cite{ding2022orientation} applied the same idea within an orientation-aware MPC. 

Despite these advances, the standard MPC+WBC pipeline faces two challenges. The first concerns constraint handling in DDP-style nonlinear MPC: most deployed formulations linearize the dynamics and simplify the contact constraints to preserve convexity, so under large disturbances the executed motion can violate friction-cone, unilateral, or ZMP conditions that the plan satisfies. Nonlinear solvers such as differential dynamic programming (DDP) and its Gauss--Newton variant, the iterative linear-quadratic regulator (iLQR)~\cite{tassa2012synthesis}, avoid this, and have been demonstrated across multi-contact DDP~\cite{mastalli2020crocoddyl}, hybrid contact-implicit iLQR-MPC~\cite{kong2023hybrid,kim2024contact}, and reduced-order-guided contact-implicit MPC for humanoids~\cite{esteban2025reduced}. DDP exploits the temporal structure of the problem, but its weakness is inequality constraints: an SQP step carries them as QP inequalities, whereas augmented Lagrangian and relaxed-barrier penalties fold them into the stage cost. Within DDP, augmented Lagrangian (AL) penalties drive feasibility through adaptive multiplier updates~\cite{jallet2022constrained}, while relaxed barriers (ReB) replace the logarithmic barrier with a quadratic extension that keeps the backward-pass Hessian bounded~\cite{grandia2019feedback}. Li and Wensing~\cite{li2020hsddp} combined the two, assigning AL to switching equality constraints and ReB to inequalities; AL-DDP has since been applied to biped gait generation with centroidal dynamics, but in an offline planning formulation~\cite{zhou2026fast}. HS-DDP itself runs offline on a 2D planar quadruped; to our knowledge, no bipedal humanoid controller has run the AL+ReB pairing inside an online receding-horizon MPC, where the narrower stability margin makes the constraint handling harder.
\IEEEpubidadjcol
The second challenge is structural: a reduced-model plan is a reference, not a feasible whole-body motion. The single-rigid-body (SRB) model omits limb momentum~\cite{dantec2024centroidal} and joint-level limits, and returns a resultant wrench per foot rather than joint torques; completing the plan in joint space is left to the execution layer. Prior work resolves this in one of three ways: tracking the MPC force references at high priority~\cite{herzog2016momentum} can produce infeasible joint commands; tracking motion references~\cite{bellicoso2018dynamic} forgoes force-level planning; and optimizing the full whole-body dynamics directly~\cite{dantec2022wholebody} remains costly. A whole-body QP that co-optimizes contact-force increments and task-space acceleration corrections reconciles the plan with the full-body dynamics under all constraints~\cite{kim2018computationally,escande2014hierarchical}. The WBC is bounded by the plan it receives: it reshapes the contact wrench and the base acceleration, but it cannot move a footstep, so any adaptation the MPC does not plan is left to the ankle and hip strategies within the support polygon the plan already fixed.

\begin{figure*}[t]
\centering
{\includegraphics[width=\textwidth]{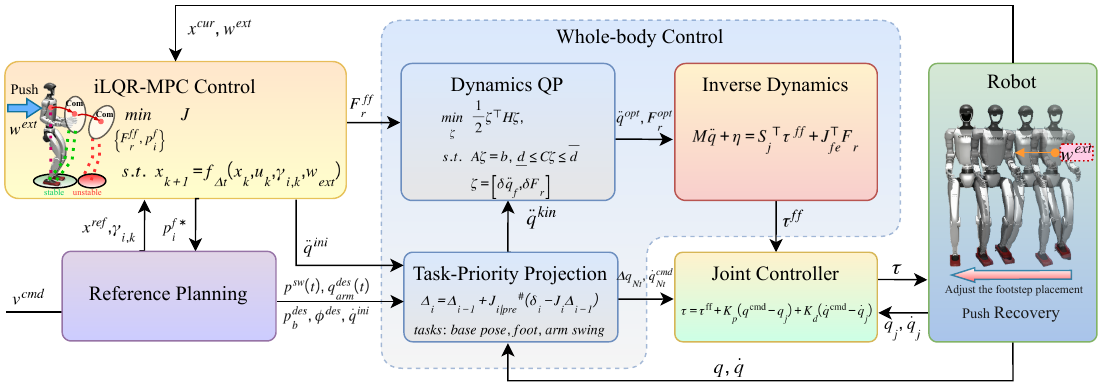}}
\caption{{Overall control architecture. A reference-planning module supplies the CoM reference, nominal footholds, and contact schedule. The reduced-model iLQR-MPC computes the contact wrench and footstep placements online from centroidal state feedback, with contact and motion constraints entering the cost $J$ as augmented-Lagrangian and relaxed-barrier penalties. The whole-body controller maps this plan to joint torques via task-priority projection, a dynamics QP, and inverse dynamics, using joint state feedback, with the QP enforcing the floating-base block of the full whole-body dynamics and contact feasibility.}}
\label{fig:control-arch}
\end{figure*}

The proposed controller addresses both challenges (Fig.~\ref{fig:control-arch}) with an iLQR-MPC and a whole-body controller; a reference-planning module upstream converts a velocity command into a CoM reference, nominal footholds, and a contact schedule. The contributions of this paper are as follows.
\begin{enumerate}
    \item \emph{Joint footstep and contact-force planning.} The controller jointly optimizes contact forces and footstep locations in a single receding-horizon iLQR-MPC on the single-rigid-body model. Both are re-optimized at every replan, so a push is answered by moving the next footstep and redistributing the stance wrench in the same solve, without a separate step planner.
    \item \emph{Online constraint handling.} Augmented-Lagrangian and relaxed-barrier penalties place the contact and motion constraints in the backward pass, which keeps the joint optimization practical at replan rate.
    \item \emph{Interpretable, model-based control.} Every stage is defined by explicit models, costs, and constraints, from the centroidal dynamics in the MPC to the null-space kinematics and floating-base QP in the whole-body layer. A disturbance response can therefore be traced to specific model terms, and no platform-specific policy training is required.
\end{enumerate}
Section~\ref{sec:wbc} details the whole-body controller, and Section~\ref{sec:experiments} reports walking and push-recovery results on the Unitree G1 in MuJoCo.

\section{Problem Formulation and \\ Theoretical Background}
\subsection{System Dynamics Modeling}
\label{sec:dynamics}

The MPC controller employs an SRB model in which contact forces and footholds are optimized jointly, and predicts the robot's motion with the continuous-time nonlinear state-space model
\begin{equation}
\label{eq:state-space-abstract}
{\dot{\bm{x}}(t) = \bm{f}\bigl(\bm{x}(t),\,\bm{u}(t);\,\bm{\gamma}(t),\,\bm{w}_{\mathrm{ext}}(t)\bigr),}
\end{equation}
{where $\bm{\gamma}=(\gamma_L,\gamma_R)$ are the gait gates from the scheduler, with $\gamma_i\in\{0,1\}$ denoting stance and swing, and $\bm{w}_{\mathrm{ext}}=[\bm{f}_{\mathrm{ext}}^\top,\bm{\tau}_{\mathrm{ext}}^\top]^\top$ is the exogenous disturbance wrench.}

Its realization consists of a state-and-disturbance term and a control-and-gait term:
\begin{equation}
\label{eq:srb-dynamics}
\resizebox{\columnwidth}{!}{%
	$
\dot{\bm{x}} =
\begin{bmatrix}
\dot{\bm{\Phi}} \\[2pt] \dot{\bm{p}} \\[2pt] \dot{\bm{\omega}} \\[2pt] \dot{\bm{v}} \\[2pt] \dot{\bm{p}}_L^f \\[2pt] \dot{\bm{p}}_R^f
\end{bmatrix}
=
\underbrace{
\begin{bmatrix}
\bm{R}_z(\psi)^\top \bm{\omega} \\[2pt]
\bm{v} \\[2pt]
\bm{I}(\bm{\Phi})^{-1}\bigl(-\bm{\omega}\times(\bm{I}(\bm{\Phi})\,\bm{\omega}) + \bm{\tau}_{\mathrm{ext}}\bigr) \\[2pt]
\frac{1}{m}\,\bm{f}_{\mathrm{ext}} + \bm{g} \\[2pt]
\bm{0}_3 \\[2pt]
\bm{0}_3
\end{bmatrix}
}_{\scriptscriptstyle f(\bm{x},\bm{w}_{\mathrm{ext}})}
\;+\;
\underbrace{
\begin{bmatrix}
\bm{0}_3 \\[2pt]
\bm{0}_3 \\[2pt]
\bm{I}(\bm{\Phi})^{-1}\!\sum\limits_{i\in\{L,R\}}\!\!\bigl(\bm{\tau}_i + {\bm{r}_i} \times \bm{F}_i\bigr) \\[6pt]
\frac{1}{m}\!\sum\limits_{i\in\{L,R\}}\!\!\bm{F}_i \\[6pt]
(1-\gamma_L)\,\bm{v}_L^f \\[4pt]
(1-\gamma_R)\,\bm{v}_R^f
\end{bmatrix}
}_{\scriptscriptstyle f(\bm{u},\bm{\gamma})}
	$
}
\end{equation}

where:
\begin{itemize}
    \item With $\bm{\Phi}=(\phi,\theta,\psi)^\top$ collecting the roll, pitch, and yaw angles, $\bm{R}_z(\psi)$ is the yaw rotation matrix; with small roll and pitch, $\dot{\bm{\Phi}}$ is well approximated by the world-frame angular velocity rotated through yaw alone.
    \item $\bm{I}(\bm{\Phi}) = \bm{R}(\bm{\Phi})\,\bar{\bm{I}}\,\bm{R}(\bm{\Phi})^\top$ is the world-frame centroidal inertia, with $\bar{\bm{I}}$ the constant body-frame inertia.
    \item $\bm{r}_i = \bm{p}_i^f - \bm{p}$ is the CoM-to-foot lever arm, and $(\bm{F}_i,\bm{\tau}_i)$ is the force and moment applied by foot $i$ to the body.
    \item $\bm{g} = [0,\,0,\,-9.81]^\top$ is gravity, grouped with the drift.
    \item $\bm{f}_{\mathrm{ext}}, \bm{\tau}_{\mathrm{ext}}$ are exogenous disturbance force and torque, used in the push-recovery analysis of Section~\ref{sec:experiments}.
\end{itemize}

The state and control vectors, partitioned in \eqref{eq:srb-dynamics}, are
\begin{equation}
\label{eq:state}
{\bm{x} = \bigl[\bm{\Phi}^\top,\,\bm{p}^\top,\,\bm{\omega}^\top,\,\bm{v}^\top,\,(\bm{p}_L^f)^\top,\,(\bm{p}_R^f)^\top\bigr]^\top \in \R^{18},}
\end{equation}
\begin{equation}
\label{eq:control}
{\bm{u} = \bigl[\bm{F}_L^\top,\,\bm{\tau}_L^\top,\,\bm{F}_R^\top,\,\bm{\tau}_R^\top,\,(\bm{v}_L^f)^\top,\,(\bm{v}_R^f)^\top\bigr]^\top \in \R^{18},}
\end{equation}
The sampling period is $\Delta t = {5}\,\mathrm{ms}$ over a horizon of $N = 10$ stages, with $\gamma_k=(\gamma_{L,k},\gamma_{R,k})$ the stage-$k$ gait gates.

\textbf{Remark.} The SRB model is a special case of centroidal dynamics~\cite{orin2013centroidal}: it fixes the angular momentum about the CoM to the rigid-body-rotation component $\bm{h}_G^\omega=\bm{I}\bm{\omega}$, neglecting limb-velocity contributions. This is standard at walking speeds~\cite{di2018dynamic}; Section~\ref{sec:wbc} restores the full floating-base dynamics in the WBC.

\subsection{iLQR Solution Framework}
\label{sec:ddp}

The MPC controller solves a receding-horizon optimal control problem that minimizes a terminal plus stage cost subject to the forward-Euler discretization of the SRB dynamics and the contact constraints of Section~\ref{sec:contact-handling}. Touch-down and lift-off are discrete mode transitions of a hybrid dynamical system~\cite{romualdi2022online}. The complete problem is \eqref{eq:E-mpc-opt}, solved via iLQR.

{Standard iLQR~\cite{tassa2012synthesis} solves this problem by expanding the action-value function $Q_k(\bm{x}_k,\bm{u}_k)=\ell(\bm{x}_k,\bm{u}_k)+V_{k+1}\big(f(\bm{x}_k,\bm{u}_k,\gamma_k)\big)$ to second order around a nominal trajectory; minimizing the quadratic model yields the feedback policy
\begin{equation}
\delta \bm{u}_k = \bm{k}_k + \bm{K}_k\,\delta\bm{x}_k,
\end{equation}
with feedforward $\bm{k}_k = -\Qddp_{\bm{u}\bm{u},k}^{-1}\Qddp_{\bm{u},k}$ and gain $\bm{K}_k = -\Qddp_{\bm{u}\bm{u},k}^{-1}\Qddp_{\bm{u}\bm{x},k}$ from the local quadratic model $\Qddp$ of the action value. The cost-to-go gradient and Hessian are propagated backward from the terminal cost, and dropping the tensor contractions of $\nabla^2 f$ gives the Gauss--Newton (iLQR) approximation. The forward pass applies $\bm{u}_k\leftarrow\hat{\bm{u}}_k+\alpha\,\bm{k}_k+\bm{K}_k(\bm{x}_k-\hat{\bm{x}}_k)$ with a line-search parameter $\alpha$.}

\section{{Constrained MPC Formulation}}
\subsection{{Contact Stability Constraints}}
\label{sec:contact-handling}

{A stable foot--ground contact requires a unilateral normal force ($F_z\ge 0$), tangential forces inside a friction pyramid so the foot does not slip, a ZMP within the support polygon, and bounded yaw torque. These define the \emph{contact stability constraints}, expressed as linear inequalities in the contact wrench and foot position and enforced by differentiable penalties (Section~\ref{sec:al-penalties}).}

\paragraph{{Support-foot wrench constraints.}}
{Let $\bm{F}_i=(F_{i,x},F_{i,y},F_{i,z})^\top$ and $\bm{\tau}_i=(\tau_{i,x},\tau_{i,y},\tau_{i,z})^\top$ denote the world-frame contact wrench of foot $i\in\{L,R\}$, given by the corresponding blocks of $\bm{u}_k$. The support region is a rectangle with longitudinal half-length $L_x$ and lateral half-width $L_y$. Within it, the wrench must satisfy unilateral contact, an inscribed friction pyramid, a no-tipping (ZMP) condition, and yaw torque bounds, summarized by}
\begin{equation}
\label{eq:E-wrench-template}
\begin{aligned}
F_{i,z}&\ge 0,\\
|F_{i,x}|,\,|F_{i,y}|&\le \rho_\mu\,F_{i,z},\\
|\tau_{i,x}|&\le L_y F_{i,z},\\
|\tau_{i,y}|&\le L_x F_{i,z},\\
\tau_{i}^{\min}\le \tau_{i,z}&\le \tau_{i}^{\max}.
\end{aligned}
\end{equation}
{Here $\rho_\mu=\mu/\sqrt{2}$ is the lateral friction bound of a four-facet pyramid inscribed in the cone of coefficient $\mu$, and $\tau_{i}^{\min},\tau_{i}^{\max}$ bound the yaw torque.}

\paragraph{{ZMP coordinates.}}
{For $F_{i,z}>0$, the horizontal ZMP coordinates follow from the moments about the footprint center in the foot frame, whose $z$ axis aligns with the contact normal:}
\begin{equation}
\label{eq:E-zmp-xy}
{x_{i}^{\mathrm{zmp}}=-\frac{\tau_{i,y}}{F_{i,z}},\qquad
y_{i}^{\mathrm{zmp}}=\frac{\tau_{i,x}}{F_{i,z}}.}
\end{equation}
Hence the torque bounds in \eqref{eq:E-wrench-template} are equivalent to $|x_{i}^{\mathrm{zmp}}|\le L_x$ and $|y_{i}^{\mathrm{zmp}}|\le L_y$, enforced through the wrench $(\bm{F}_i,\bm{\tau}_i)$ without introducing the ZMP coordinates as decision variables.

\paragraph{{Contact location box constraints.}}
In swing, the swing-foot position $\bm{p}_{\mathrm{sw}}^f$ is held inside a box of half-extents $\bm{\ell}=(\ell_x,\ell_y,\ell_z)^\top>0$ centered on the nominal foothold $\bm{p}_{\mathrm{sw}}^{\mathrm{nom}}$, whose horizontal components come from the gait scheduler and whose vertical component is fixed at $z_{\mathrm{ref}}$:
\begin{equation}
\label{eq:E-contact-location-box}
{-\bm{\ell}\;\le\;\bm{p}_{\mathrm{sw}}^f-\bm{p}_{\mathrm{sw}}^{\mathrm{nom}}\;\le\;\bm{\ell},}
\end{equation}
interpreted componentwise with $\ell_z=0$. When step adjustment is active, $\bm{p}_{\mathrm{sw}}^{\mathrm{nom}}$ is updated online from the previous MPC solution; otherwise it stays at the open-loop gait-scheduler value. Each component yields a nonnegative slack for the ReB penalty (Section~\ref{sec:al-penalties}).

\subsection{{Penalty-Based Constraint Enforcement}}
\label{sec:al-penalties}

The iLQR solver assumes a smooth, twice-differentiable objective, so the hard constraints of Section~\ref{sec:contact-handling}, which define the stage-wise admissible state and control sets $\mathcal{X}_k$ and $\mathcal{U}_k$, are converted into additive penalty terms: augmented Lagrangian (AL) penalties for the swing-foot zero-wrench equality, and relaxed logarithmic barriers (ReB) for the friction, ZMP, unilateral, yaw-torque, and foothold-box inequalities. Both enter the iLQR backward pass through analytic gradients and Hessians, so iLQR handles the constraints without explicit projection. The total cost in \eqref{eq:E-mpc-opt} collects the stage-tracking, control-rate, and terminal terms together with $J_{\mathrm{AL}}$ and $J_{\mathrm{ReB}}$ (Section~\ref{sec:complete-mpc}).

\paragraph{{Augmented Lagrangian for the swing-foot zero-wrench equality.}}
In single support the swing foot must apply zero wrench,
\begin{equation}
\label{eq:E-zeta-sw}
\bm{u}_k^{\mathrm{sw}}
=
\begin{bmatrix} \bm{F}_{\mathrm{sw}} \\ \bm{\tau}_{\mathrm{sw}} \end{bmatrix}\in\R^6,
\end{equation}
so the equality constraint is
\begin{equation}
\label{eq:E-h-eq}
\bm{h}_k(\bm{u}_k)=\bm{u}_k^{\mathrm{sw}}=\bm{0}.
\end{equation}
Let $i_{\mathrm{sw}}\in\{L,R\}$ be the swing foot; $\bm{h}_k$ selects $(\bm{F}_{i_{\mathrm{sw}}},\bm{\tau}_{i_{\mathrm{sw}}})$ from $\bm{u}_k$. In double support, which occurs only when standing still, there is no swing foot and the constraint is dropped ($\bm{h}_k\equiv\bm{0}$). The augmented-Lagrangian term in the total cost is
\begin{equation}
\label{eq:E-al-total}
J_{\mathrm{AL}}
=
\sum_{k=0}^{N-1}
\mathbf{1}_{\mathrm{SS},k}
\left(
\bm{\lambda}_k^\top \bm{h}_k(\bm{u}_k)
+
\frac{\sigma}{2}\norm{\bm{h}_k(\bm{u}_k)}^2
\right),
\end{equation}
{where the single-support indicator $\mathbf{1}_{\mathrm{SS},k}\in\{0,1\}$ restricts the penalty to single-support phases, $\bm{\lambda}_k\in\R^6$ are dual variables, and $\sigma>0$ is the penalty parameter. Because $\bm{h}_k$ is affine in the control, $\bm{h}_k=\bm{S}_k\bm{u}_k$ with a constant row-selector $\bm{S}_k$, its gradient, Hessian, and dual update are explicit:}
\begin{equation}
\label{eq:E-al-derivs}
\resizebox{\columnwidth}{!}{%
$
\nabla_{\bm{u}}J_{\mathrm{AL}}=\bm{S}_k^\top(\bm{\lambda}_k+\sigma\bm{h}_k),\quad
\nabla^2_{\bm{u}\bm{u}}J_{\mathrm{AL}}=\sigma\,\bm{S}_k^\top\bm{S}_k,\quad
\bm{\lambda}_k\leftarrow\bm{\lambda}_k+\sigma\,\bm{h}_k
$}
\end{equation}
{contributing only to the control channel ($\nabla_{\bm{x}}J_{\mathrm{AL}}=\bm{0}$).}

\paragraph{{Relaxed barrier for inequality constraints.}}
{Each scalar inequality is expressed as a nonnegative slack $c_{j,k}(\bm{x}_k,\bm{u}_k)>0$ and penalized by the \emph{relaxed} logarithmic barrier~\cite{grandia2019feedback}, with argument $z=c_{j,k}$, smoothing parameter $\delta>0$, polynomial degree parameter $p=2$, and barrier weight $\varepsilon_B>0$:}
\begin{equation}
\label{eq:E-reb-barrier}
B_\delta(z)=
\begin{cases}
-\ln z, & z>\delta,\\[0.5ex]
\displaystyle\frac{p-1}{p}\left[\left(\frac{z-p\delta}{(p-1)\delta}\right)^{\!p}{-1}\right]-\ln\delta, & z\le \delta.
\end{cases}
\end{equation}
For $z>\delta$ this is the logarithmic barrier; the polynomial branch replaces it on $z\le\delta$, keeping $B_\delta$ and its first two derivatives finite and conditioning the backward pass. Its derivatives are
\begin{equation}
\label{eq:E-reb-derivs}
\resizebox{\columnwidth}{!}{%
$
B_\delta'(z)=
\begin{cases}
-1/z, & z>\delta,\\[0.3ex]
\frac{1}{\delta}\bigl(\tfrac{z-p\delta}{(p-1)\delta}\bigr)^{p-1}, & z\le\delta,
\end{cases}
\qquad
B_\delta''(z)=
\begin{cases}
1/z^{2}, & z>\delta,\\[0.3ex]
\frac{1}{\delta^{2}}\bigl(\tfrac{z-p\delta}{(p-1)\delta}\bigr)^{p-2}, & z\le\delta,
\end{cases}
$}
\end{equation}
{which enter the backward pass as $\nabla_{\bm{u}}J_{\mathrm{ReB}}\mathrel{+}=\varepsilon_B\,B_\delta'(c_{j,k})\nabla_{\bm{u}}c_{j,k}$ and a Gauss--Newton Hessian $\varepsilon_B\,B_\delta''(c_{j,k})(\nabla_{\bm{u}}c_{j,k})(\nabla_{\bm{u}}c_{j,k})^\top$ per slack, with the barrier parameter tightened as $\delta\leftarrow\max\{\eta_\delta\delta,\delta_{\min}\}$ ($\eta_\delta\in(0,1)$, $\delta_{\min}>0$).} The total ReB objective is
\begin{equation}
\label{eq:E-reb-total}
J_{\mathrm{ReB}}
=
\varepsilon_B\sum_{k=0}^{N-1}\sum_{j}B_\delta\big(c_{j,k}(\bm{x}_k,\bm{u}_k)\big),
\end{equation}
{the inner sum runs over all friction-pyramid, unilateral-contact, ZMP, yaw-torque, and foothold-box slacks of Section~\ref{sec:contact-handling}.}

\subsection{{Disturbance Rejection via Receding-Horizon Replanning}}

\paragraph{{Footstep-in-state formulation.}}
{Re-solving the finite-horizon problem each cycle adapts contact forces and CoM motion to disturbances without a hand-designed compensator~\cite{wieber2006trajectory, kuindersma2016optimization}. Footstep locations are promoted to state variables, and the foot positions $\bm{p}_i^f\;(i\in\{L,R\})$ evolve with the commanded foot velocities $\bm{v}_i^f$ under the gait gates $\gamma_{i,k}$~\cite{romualdi2022online}:}
\begin{equation}
\label{eq:E-foot-discrete}
{\bm{p}_{i,k+1}^f=\bm{p}_{i,k}^f+\Delta t\,(1-\gamma_{i,k})\,\bm{v}_{i,k}^f,\qquad i\in\{L,R\}.}
\end{equation}
{The contact-location box on these foot positions enters the backward pass of the nonlinear iLQR-MPC as a relaxed-barrier term (Section~\ref{sec:al-penalties}), so the foothold is optimized in the same solve as the contact wrench.} A stance foot ($\gamma_i=1$) stays fixed; a swing foot ($\gamma_i=0$) advances toward a target set by a heuristic rule on desired velocity and swing phase during nominal walking, or by the MPC-optimized horizontal foothold under a disturbance.

\subsection{{Complete MPC Problem}}
\label{sec:complete-mpc}

{The complete finite-horizon problem solved at each MPC cycle is:}
\begin{equation}
\label{eq:E-mpc-opt}
\footnotesize
\begin{aligned}
\min_{\{\bm{u}_k\}_{k=0}^{N-1}}
&\sum_{k=0}^{N-1}
\ell(\bm{x}_k,\bm{u}_k)
+
\rho_{\Delta} \sum_{k=1}^{N-1} \lVert \bm{u}_k-\bm{u}_{k-1}\rVert^2
\\
&\qquad +
\eta_N\,\ell_N(\bm{x}_N)
+
J_{\mathrm{AL}}\big(\{\bm{u}_k\}\big)
+
J_{\mathrm{ReB}}\big(\{\bm{x}_k,\bm{u}_k\}\big), \\
\text{s.t.}\quad
&\bm{x}_{k+1}=f_{\Delta t}\!\left(\bm{x}_k,\bm{u}_k;\gamma_k, \bm{w}_{\mathrm{ext}}\right),\
k=0,\ldots,N-1,\\
&\bm{p}_{i,k+1}^f=\bm{p}_{i,k}^f+\Delta t\,(1-\gamma_{i,k})\,\bm{v}_{i,k}^f,\
{\scriptstyle i\in\{L,R\},\ k=0,\ldots,N-1}.
\end{aligned}
\end{equation}
{The stage cost $\ell$ collects quadratic penalties on attitude, CoM position, velocities, contact wrench, angular momentum, foot placement, and footstep velocity; a control-rate term $\rho_{\Delta}$ {penalizes consecutive command differences}, and a terminal cost $\eta_N$ regularizes the final state. The penalties \eqref{eq:E-al-total} and \eqref{eq:E-reb-total} are defined in Section~\ref{sec:al-penalties}.}

{The constraints are the numerically integrated SRB dynamics \eqref{eq:srb-dynamics} and the gated footstep recursion \eqref{eq:E-foot-discrete}.} {Each cycle, the re-measured state $\bm{x}^{\mathrm{cur}}$ initializes the problem and the disturbance wrench $\bm{w}_{\mathrm{ext}}$ is refreshed. In the simulations, $\bm{w}_{\mathrm{ext}}$ is treated as known.}

\section{Whole-Body Motion Controller}
\label{sec:wbc}

{The MPC controller (Sections~\ref{sec:dynamics}--\ref{sec:complete-mpc}) produces a centroidal trajectory, the contact wrench, and the optimal footstep locations $\bm{p}_i^{f\star}\;(i\in\{L,R\})$. The whole-body controller (WBC) maps this plan to joint torques while respecting the full floating-base dynamics and contact constraints. The WBC implements a hierarchical task-priority architecture~\cite{sentis2005control} with three stages: a kinematic null-space recursion (Section~\ref{sec:wbc-kin}) emitting nominal motion commands, a compact quadratic program (Section~\ref{sec:wbc-qp}) correcting the floating-base acceleration and contact wrench under the equations of motion and contact constraints, and an inverse-dynamics step (Section~\ref{sec:wbc-tau}) resolving the corrected quantities to joint torques. Only the last two stages produce forces and feedforward torques.}

\subsection{{Floating-Base Dynamics Model}}
\label{sec:wbc-dynamics}

{The floating-base dynamics are governed by the equations of motion}
\begin{equation}
\label{eq:floating-base-eom}
{
\bm{M}(\bm{q})\ddot{\bm{q}} + \bm{\eta}(\bm{q},\dot{\bm{q}}) = \bm{S}_j^\top\bm{\tau}^{\mathrm{ff}} + \bm{J}_{fe}^\top(\bm{q})\,\bm{F}_r,
}
\end{equation}
{where $\bm{q}=[\bm{q}_f^\top,\bm{q}_j^\top]^\top\in\R^{n_q}$ are the generalized coordinates, with floating base $\bm{q}_f\in\R^6$ and actuated joints $\bm{q}_j\in\R^{n_j}$, so $n_q=6+n_j=35$; $\bm{M}\in\R^{n_q\times n_q}$ is the positive-definite mass matrix, $\bm{\eta}$ collects Coriolis and gravity terms, $\bm{S}_j\in\R^{n_j\times n_q}$ selects the actuated joint rows, $\bm{J}_{fe}\in\R^{12\times n_q}$ is the contact Jacobian, $\bm{F}_r\in\R^{12}$ is the contact wrench, and $\bm{\tau}^{\mathrm{ff}}\in\R^{n_j}$ are the feedforward joint torques.}

\subsection{{Null-space task prioritization}}
\label{sec:wbc-kin}
Each task is solved in the null space of all higher-priority tasks~\cite{kim2019wbic}, so lower-priority corrections cannot disturb higher-priority constraints.

\paragraph{{Null-space propagation.}}
{Let $\bm{N}_{i-1}$ be the cumulative null space of all tasks preceding the $i$-th and let $N_t$ be the number of tasks. The projected Jacobian is $\bm{J}_{i|\mathrm{pre}}=\bm{J}_i(\bm{q})\bm{N}_{i-1}$, and the elementary projector built from it uses the Moore--Penrose inverse $\{\cdot\}^\dagger$ in the position and velocity passes and the dynamically consistent inverse $\overline{(\cdot)}=\bm{M}^{-1}(\cdot)^\top\bigl((\cdot)\bm{M}^{-1}(\cdot)^\top\bigr)^{-1}$ in the acceleration pass.}

\paragraph{{Task-space recursion.}}
{At priority level $i$ the task command is projected into this null space and accumulated onto the preceding level. With the PD task law $\ddot{\bm{x}}_i^{\mathrm{cmd}}=\ddot{\bm{x}}_i^{\mathrm{des}}+\bm{K}_{p,i}(\bm{x}_i^{\mathrm{des}}-\bm{x}_i)+\bm{K}_{d,i}(\dot{\bm{x}}_i^{\mathrm{des}}-\dot{\bm{x}}_i)$ and task error $\bm{e}_i=\bm{x}_i^{\mathrm{des}}-\bm{x}_i$, the position, velocity, and acceleration levels share one recursion}
\begin{equation}
\label{eq:E-wbc-recursion}
{\bm{\Delta}_i=\bm{\Delta}_{i-1}+\bm{J}_{i|\mathrm{pre}}^{\#}\bigl(\bm{\delta}_i-\bm{J}_i\bm{\Delta}_{i-1}\bigr),\qquad i=1,\dots,N_t,}
\end{equation}
{with $(\bm{\Delta}_i,\bm{\delta}_i,\bm{J}_{i|\mathrm{pre}}^{\#})$ equal to $(\Delta\bm{q}_i,\bm{e}_i,\bm{J}_{i|\mathrm{pre}}^\dagger)$ at the position level, $(\dot{\bm{q}}_i^{\mathrm{cmd}},\dot{\bm{x}}_i^{\mathrm{des}},\bm{J}_{i|\mathrm{pre}}^\dagger)$ at the velocity level, and $(\ddot{\bm{q}}_i^{\mathrm{cmd}},\ddot{\bm{x}}_i^{\mathrm{cmd}}-\dot{\bm{J}}_i\dot{\bm{q}},\overline{\bm{J}}_{i|\mathrm{pre}})$ at the acceleration level. The final acceleration $\ddot{\bm{q}}^{\mathrm{kin}}=\ddot{\bm{q}}_{N_t}^{\mathrm{cmd}}$ resolves all task conflicts kinematically and serves as the linearization point for the QP below; the joint increment $\Delta\bm{q}_{N_t}$ and velocity $\dot{\bm{q}}_{N_t}^{\mathrm{cmd}}$ form the joint kinematic commands.}

The task-space references are the desired base pose $(\bm{p}_b^{\mathrm{des}},\bm{\Phi}^{\mathrm{des}})$, the desired foot trajectory $\bm{p}^{sw}(t)$, and the desired arm configuration $\bm{q}_{\mathrm{arm}}^{\mathrm{des}}(t)$, with the last two supplied by the gait and arm planners. The initial generalized velocity $\dot{\bm{q}}^{\mathrm{ini}}$ holds the commanded planar and yaw velocities, while the initial acceleration $\ddot{\bm{q}}^{\mathrm{ini}}$ takes its horizontal component from the MPC contact forces through the SRB dynamics and its yaw component from a yaw-rate correction; the tasks fill the remaining entries.

\subsection{QP Formulation}
\label{sec:wbc-qp}

{The decision variable $\bm{\zeta}=[\delta\ddot{\bm{q}}_{{f}}^\top,\,\delta\bm{F}_r^\top]^\top\in\R^{18}$ stacks a correction to the floating-base acceleration ($\delta\ddot{\bm{q}}_{{f}}\in\R^6$) and to the contact wrench ($\delta\bm{F}_r\in\R^{12}$). Let $\bm{F}_r^{\mathrm{ff}}\in\R^{12}$ denote the MPC feedforward contact wrench and $\ddot{\bm{q}}^{\mathrm{kin}}$ the {nominal whole-body} acceleration from the prioritized task recursion of Section~\ref{sec:wbc-kin}. The QP minimizes a weighted least-squares objective subject to the linearized floating-base dynamics and contact constraints:}
\begin{equation}
\label{eq:E-wbc-qp}
{
\begin{aligned}
\min_{\bm{\zeta}}\quad
& \tfrac{1}{2}\bm{\zeta}^\top \bm{H}\bm{\zeta},
\\
\text{s.t.}\quad
& \bm{A}\bm{\zeta}=\bm{b},
\qquad
\underline{\bm{d}}\le \bm{C}\bm{\zeta}\le \overline{\bm{d}}.
\end{aligned}}
\end{equation}
{The equality constraint enforces the floating-base block of the full dynamics; the inequalities enforce contact feasibility on the stance feet and a zero wrench on the swing foot, with $\bm{C}$ and the bounds collecting the linearized limits. Since joint accelerations remain at their task-recursion values, only the base acceleration and contact wrench are corrected. Linearizing about $(\ddot{\bm{q}}^{\mathrm{kin}},\bm{F}_r^{\mathrm{ff}})$ gives $\bm{A}\bm{\zeta}=\bm{b}$, with $\bm{A}=[\bm{S}_f\bm{M}\bm{S}_f^\top,\,-\bm{S}_f\bm{J}_{fe}^\top]$ and $\bm{b}=\bm{S}_f(\bm{J}_{fe}^\top\bm{F}_r^{\mathrm{ff}}-\bm{M}\ddot{\bm{q}}^{\mathrm{kin}}-\bm{\eta})$, and $\bm{S}_f=[\bm{I}_6\;\bm{0}]$ extracts the floating-base rows. The weight $\bm{H}\succ 0$ keeps the base acceleration near its task-recursion value, so the contact forces absorb the residual. With the optimal correction $\bm{\zeta}^\star$, the resolved acceleration and contact forces are}
\begin{equation}
\label{eq:E-wbc-post-qp}
{\ddot{\bm{q}}^{\mathrm{opt}}=\ddot{\bm{q}}^{\mathrm{kin}}+\bm{S}_f^\top\delta\ddot{\bm{q}}_{{f}}^\star,\qquad
\bm{F}_r^{\mathrm{opt}}=\bm{F}_r^{\mathrm{ff}}+\delta\bm{F}_r^\star,}
\end{equation}

\subsection{{Joint torque computation}}
\label{sec:wbc-tau}
{Given the QP-adjusted acceleration $\ddot{\bm{q}}^{\mathrm{opt}}$ and contact wrench $\bm{F}_r^{\mathrm{opt}}$ assembled in \eqref{eq:E-wbc-post-qp}, the feedforward joint torques follow from the floating-base inverse dynamics:}
\begin{equation}
\label{eq:E-wbc-tau}
\bm{\tau}^{\mathrm{ff}}=\bm{S}_j\bigl(\bm{M}\ddot{\bm{q}}^{\mathrm{opt}}+\bm{\eta}-\bm{J}_{fe}^\top\bm{F}_r^{\mathrm{opt}}\bigr),
\end{equation}
{where $\bm{M}$, $\bm{\eta}$, $\bm{S}_j$, and $\bm{J}_{fe}$ are as defined in \eqref{eq:floating-base-eom}, and $\bm{\tau}^{\mathrm{ff}}$ is the feedforward joint torque, distinct from the foot contact moments $\bm{\tau}_L,\bm{\tau}_R$ in the SRB model. These torques are combined with joint PD feedback to form the motor command,
\begin{equation}
\label{eq:E-joint-pd}
\bm{\tau}=\bm{\tau}^{\mathrm{ff}}+\bm{K}_p(\bm{q}^{\mathrm{cmd}}-\bm{q}_j)+\bm{K}_d(\dot{\bm{q}}^{\mathrm{cmd}}-\dot{\bm{q}}_j),
\end{equation}
{where $\bm{q}^{\mathrm{cmd}}$ and $\dot{\bm{q}}^{\mathrm{cmd}}$ are the joint-space position and velocity commands, formed from the joint blocks of $\Delta\bm{q}_{N_t}$ and $\dot{\bm{q}}_{N_t}^{\mathrm{cmd}}$, $\bm{q}_j$ and $\dot{\bm{q}}_j$ are the measured joint positions and velocities, and $\bm{K}_p,\bm{K}_d$ are the joint PD gains.}}

\section{Simulation Experiments}
\label{sec:experiments}

\subsection{Experimental Setup}
\label{sec:setup}

All experiments use the MuJoCo simulator~\cite{todorov2012mujoco} and a Unitree G1 model. Simulation and the WBC (Section~\ref{sec:wbc}) run synchronously at $500\,\mathrm{Hz}$; ground contact uses a penalty-based soft-constraint model.

The iLQR-MPC uses $N=10$ and $\Delta t={5}\,\mathrm{ms}$ over a ${50}\,\mathrm{ms}$ horizon, is replanned every $10\,\mathrm{ms}$ with zero-order hold, and takes $\sim{9}\,\mathrm{ms}$ per solve. It is implemented in CasADi~\cite{andersson2019casadi} with its hot path accelerated in native C++; the WBC QP is solved with qpOASES~\cite{ferreau2014qpoases}. Experiments run on an Intel Core i5-12400F CPU with 16\,GB RAM.

\subsection{Steady-State Locomotion Performance}
\label{sec:steady}

We first characterize baseline tracking during unperturbed forward walking on flat ground at $0.1\,\mathrm{m/s}$ for approximately $40\,\mathrm{s}$, comparing the closed-loop CoM state $\bm{x}^{\mathrm{cur}}$ against the MPC reference $\bm{x}^{\mathrm{ref}}$, both recorded at $500\,\mathrm{Hz}$.

\begin{figure}[H]
\centering
\includegraphics[width=\columnwidth]{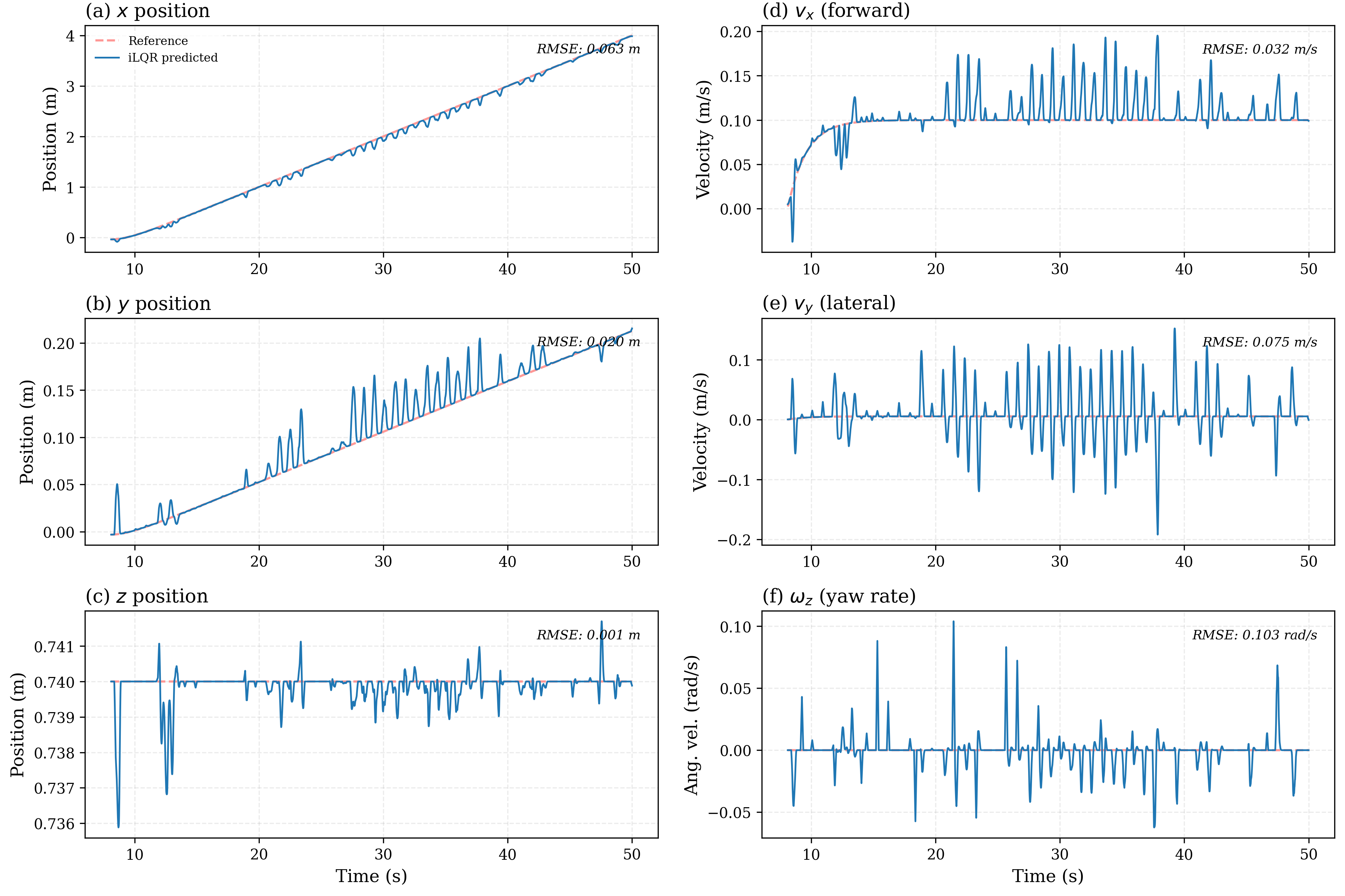}
\caption{{MPC horizon prediction of the CoM state. (a)--(c) CoM position; (d)--(f) planar velocity and yaw rate. The reference $\bm{x}^{\mathrm{ref}}$ (light red, dashed) and prediction $\bm{x}^{\mathrm{pred}}$ (blue) are shown; RMSE is annotated per panel.}}
\label{fig:mpc-prediction}
\end{figure}

{Fig.~\ref{fig:mpc-prediction} shows the MPC prediction against the reference. Prediction tracked the reference closely; the accumulated residual was largest in the forward position, and footstep-induced oscillations are visible in the lateral velocity and yaw rate (panels~(e) and~(f)).}

\begin{table}[H]
\centering
\caption{{Per-State Tracking RMSE During 40\,s Forward Walking at $0.1\,\mathrm{m/s}$}}
\label{tab:tracking-rmse}
\begin{tabular}{lclc}
\toprule
State & RMSE & State & RMSE \\
\midrule
Roll (rad) & 0.038 & $\omega_x$ (rad/s) & 0.214 \\
Pitch (rad) & 0.016 & $\omega_y$ (rad/s) & 0.114 \\
Yaw (rad) & 0.029 & $\omega_z$ (rad/s) & 0.175 \\
$X$ (m) & 0.231 & $v_x$ (m/s) & 0.078 \\
$Y$ (m) & 0.064 & $v_y$ (m/s) & 0.267 \\
$Z$ (m) & 0.004 & $v_z$ (m/s) & 0.064 \\
\bottomrule
\end{tabular}
\end{table}

Table~\ref{tab:tracking-rmse} reports the closed-loop tracking performance over the same segment. The largest error is the forward CoM position ($0.231\,\mathrm{m}$), dominated by the startup transient; attitude and the remaining states stay bounded.

\subsection{Disturbance Rejection}
\label{sec:disturbance}

To assess robustness, impulsive external forces are applied to the torso during forward walking at $0.1\,\mathrm{m/s}$: a $0.12\,\mathrm{s}$ force pulse at $t = 15.0\,\mathrm{s}$, independently along $+x$, $-x$, $+y$, and $-y$, with the magnitude increased from $80\,\mathrm{N}$ to the failure threshold, defined as any non-foot link contacting the ground. The controller also recovers from oblique pushes that superimpose sagittal and lateral components.

\subsubsection{Sagittal push recovery}

\begin{figure*}[!b]
\centering
\includegraphics[width=\textwidth]{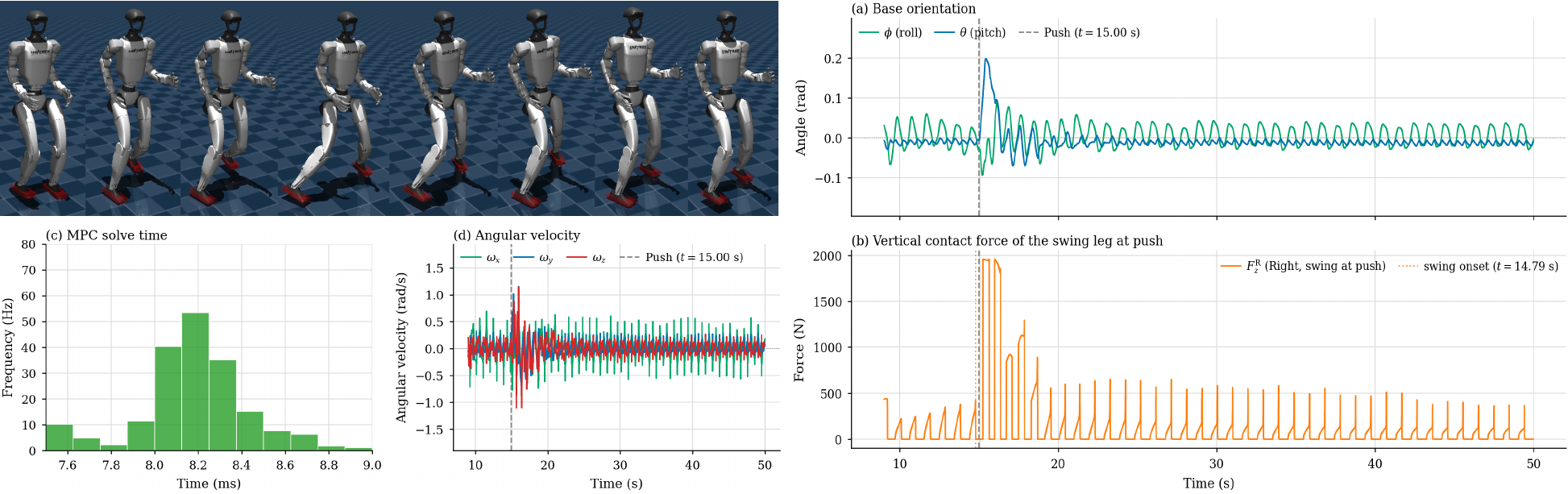}
\caption{\rev{Forward ($+x$) push recovery at $245\,\mathrm{N}$ at $15.0\,\mathrm{s}$ during $0.1\,\mathrm{m/s}$ walking. Left: walking sequence; (c) iLQR solve-time distribution. Right: (a) base orientation; (b) swing-leg vertical contact force; (d) angular velocity.}}
\label{fig:push-recovery-plusx}
\end{figure*}

{Table~\ref{tab:push-recovery} reports forward ($+x$) pushes sustained up to $245\,\mathrm{N}$; beyond this force, we attribute the failure to the MPC's limited capacity to decelerate the body within the available stepping horizon. At $245\,\mathrm{N}$, the torso pitch $\theta$ rises \rev{to about $0.20\,\mathrm{rad}$} while the roll $\phi$ \rev{stays within about $\pm 0.10\,\mathrm{rad}$} as the MPC adjusts the footstep plan to arrest the displacement. Angular velocity \rev{peaks near $1.1\,\mathrm{rad/s}$} and is suppressed within $1.0\,\mathrm{s}$; \rev{the swing-leg vertical force spikes to about $2000\,\mathrm{N}$ at push onset, consistent with the stepping response.}}

\subsubsection{Lateral push recovery}

Table~\ref{tab:push-recovery} reports the lateral outcomes, which are strongly asymmetric: left-to-right ($-y$) pushes are tolerated up to $175\,\mathrm{N}$, whereas right-to-left ($+y$) pushes fail above $80\,\mathrm{N}$, more than a twofold difference. This asymmetry arises from the stance-leg phase at push onset. With the left leg in stance, a left-to-right ($-y$) push is directed toward the stance leg, where the contact wrench redistributes directly to counteract it. Conversely, a right-to-left ($+y$) push drives the CoM away from the stance foot, shrinking the support polygon and leaving footstep adjustment as the sole recovery mechanism; the narrow mediolateral stance further limits the feasible footstep displacement.

\begin{table}[!t]
\centering
\caption{Maximum Recoverable Push Force During Forward Walking at $0.1\,\mathrm{m/s}$}
\label{tab:push-recovery}
\begin{tabular}{lcc}
\toprule
Direction & Type & Max Force (N) \\
\midrule
$+x$ (forward)         & sagittal & 245 \\
$-x$ (backward)        & sagittal & 205 \\
$+y$ (right-to-left)   & lateral  & 80  \\
$-y$ (left-to-right)   & lateral  & 175 \\
\bottomrule
\end{tabular}
\end{table}

\begin{figure*}[!b]
\centering
\includegraphics[width=\textwidth]{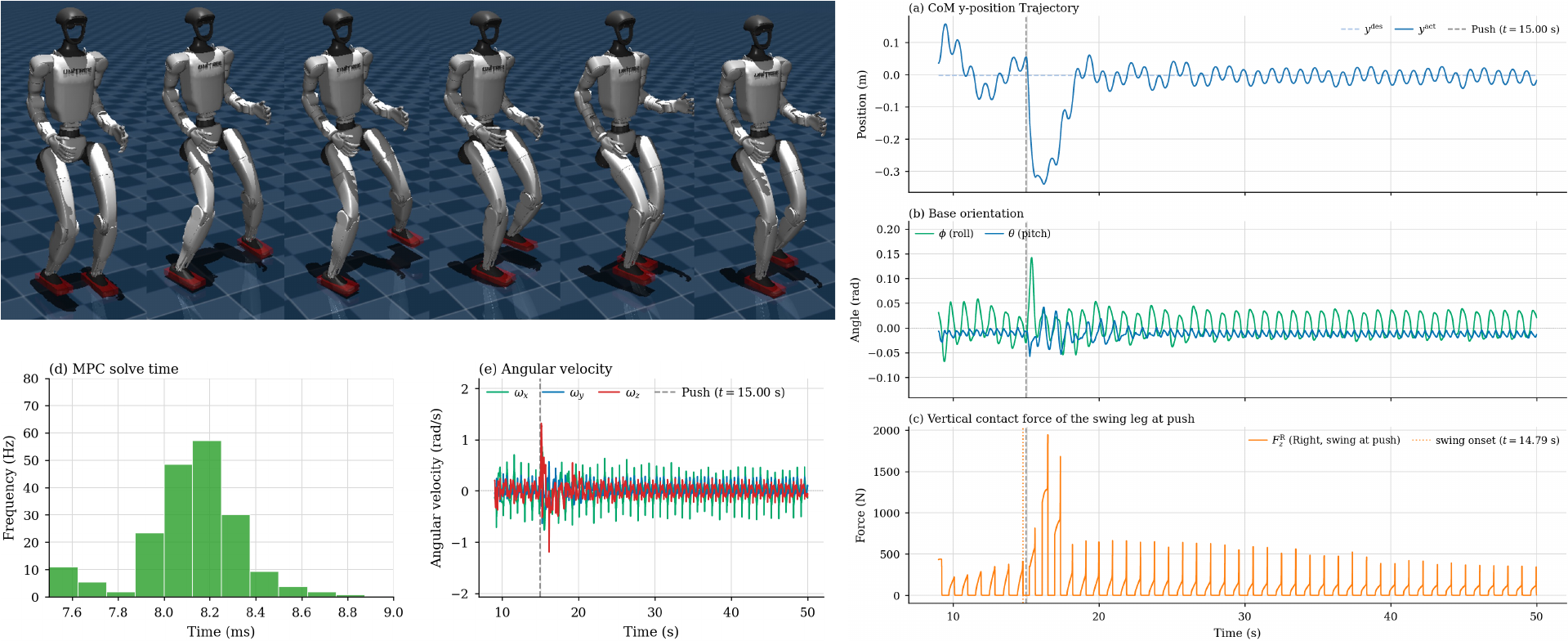}
\caption{\rev{Left-to-right ($-y$) push recovery at $175\,\mathrm{N}$ at $15.0\,\mathrm{s}$ during $0.1\,\mathrm{m/s}$ walking. Left: walking sequence; (d) iLQR solve-time distribution. Right: (a) CoM lateral position; (b) base orientation; (c) swing-leg vertical contact force; (e) angular velocity.}}
\label{fig:push-recovery-minusy}
\end{figure*}

Fig.~\ref{fig:push-recovery-minusy} summarises the response to the favorable ($-y$) lateral push, and the unfavorable ($+y$) direction is reported in Table~\ref{tab:push-recovery}. In contrast to the sagittal case, lateral pushes primarily excite $v_y$ and the roll mode, while $v_x$ remains largely unperturbed.

{Under a right-to-left ($+y$) push of $80\,\mathrm{N}$, the unfavorable direction, the disturbance acts away from the stance leg: the CoM is displaced by \rev{about $0.13\,\mathrm{m}$} before the MPC adjusts footstep placement, and the torso roll reaches \rev{about $0.06\,\mathrm{rad}$}, regulated back within $\sim$$1.0\,\mathrm{s}$. A left-to-right ($-y$) push of $175\,\mathrm{N}$ instead acts toward the stance leg: the CoM displacement grows to \rev{about $0.34\,\mathrm{m}$} with recovery within $1.0$--$1.5\,\mathrm{s}$, the roll reaches \rev{about $0.14\,\mathrm{rad}$}, and the swing-leg vertical force spikes to \rev{about $1900\,\mathrm{N}$} at touchdown.}

\subsection{Computational Performance}
\label{sec:timing}

{Table~\ref{tab:timing} breaks down the computation over $\sim$9000 iLQR-MPC solves. Each solve comprises three AL/ReB outer iterations, so the $\sim$9000 solves correspond to 27{,}000 outer iterations. One inner iLQR iteration costs $2.24\,\mathrm{ms}$, dominated by cost quadratization ($54\%$); summed over all iterations, the iLQR solve averages $8.26\,\mathrm{ms}$, and reference preparation and result conversion add $0.67\,\mathrm{ms}$, for $8.93\,\mathrm{ms}$ end to end. Across push directions the iLQR solve mean varied between $8.19$ and $8.32\,\mathrm{ms}$, and individual solves reached $13.56\,\mathrm{ms}$ during push transients.}

\begin{table}[!t]
\centering
\caption{Computation Time per Iteration and per iLQR-MPC Solve}
\label{tab:timing}
\begin{tabular}{lcc}
\toprule
Module & Time (ms) & Share (\%) \\
\midrule
\multicolumn{3}{c}{\textit{iLQR inner loop (per iteration, $N=10$)}} \\
\quad Dynamics linearization & 0.083 & 3.7 \\
\quad Cost quadratization & 1.204 & 53.8 \\
\quad Backward pass \& gain computation & 0.274 & 12.2 \\
\quad Forward rollout & 0.680 & 30.3 \\
\quad \textbf{iLQR iteration total} & \textbf{2.24} & 100 \\
\midrule
\multicolumn{3}{c}{\textit{AL-constrained MPC (per solve)}} \\
\quad iLQR solve (all AL/ReB rounds) & 8.26 & 92.4 \\
\quad Reference, initialization, conversion & 0.67 & 7.6 \\
\quad \textbf{MPC total} & \textbf{8.93} & 100 \\
\midrule
\multicolumn{3}{c}{\textit{Whole-body controller} (500\,Hz)} \\
\quad Kinematic null-space recursion & 0.136 & --- \\
\quad WBC-QP & 0.046 & --- \\
\quad Inverse dynamics & 0.001 & --- \\
\quad \textbf{WBC total} & \textbf{0.183} & --- \\
\bottomrule
\end{tabular}
\end{table}

\section{{Conclusion}}
{
{We presented a controller for robust humanoid locomotion, coupling iLQR-MPC with whole-body control. Its central advance is deciding contact forces and footstep locations in the same receding-horizon MPC, with the AL/ReB pairing running inside that loop on a torque-controlled humanoid, and every stage defined by explicit models and constraints. In MuJoCo, the Unitree G1 sustained steady-state walking and recovered from a $245\,\mathrm{N}$ forward push and a $175\,\mathrm{N}$ lateral push, with contact forces and footstep locations adapted online and a mean solve of $8.9\,\mathrm{ms}$; the lateral asymmetry arose from the stance-leg side at push onset. Three limitations remain: the walking speed is low, MPC solve times occasionally exceed the control period during push transients, and the disturbance wrench is assumed known rather than measured. Future work will focus on increasing walking speed, integrating a learned policy for disturbance perception, adding perception-driven footstep planning, and validating on hardware.}
}

\vfill

\end{document}